\documentclass[sigconf]{acmart}

\usepackage{subcaption}
\usepackage{tabularx} 

\usepackage{tikz}
\usetikzlibrary{matrix, positioning, fit, arrows.meta, backgrounds}
\usepackage{fontawesome5} % Required for the vector icons
\usepackage{caption}
\usepackage{placeins}
\usepackage{xcolor,colortbl}         % colors
\usepackage{mathtools}
\usepackage{graphicx}
\usepackage{caption}
\usepackage{subcaption}
\usepackage{algorithm}
\usepackage{algpseudocode}
\usepackage{setspace}
\usepackage{amsthm,amsfonts}
\usepackage[most]{tcolorbox}
\usepackage{fvextra}
\usepackage[T1]{fontenc}
\usepackage{enumitem}

\AtBeginDocument{%
  }

\setcopyright{acmlicensed}
\copyrightyear{2018}
\acmYear{2018}
\acmDOI{XXXXXXX.XXXXXXX}
\acmConference[Conference acronym 'XX]{Make sure to enter the correct
  conference title from your rights confirmation email}{June 03--05,
  2018}{Woodstock, NY}

\acmISBN{978-1-4503-XXXX-X/2018/06}

\copyrightyear{2026}
\acmYear{2026}
\setcopyright{cc}
\setcctype{by}
\acmConference[HCOMP 2026]{2026 ACM Conference on Human-AI Complementarity and Alignment}{September 27--30, 2026}{Alexandria, VA, USA}
\acmBooktitle{2026 ACM Conference on Human-AI Complementarity and Alignment (HCOMP 2026), September 27--30, 2026, Alexandria, VA, USA}
\acmDOI{10.1145/3834580.3838741}
\acmISBN{979-8-4007-2894-5/2026/09}

\begin{document}

\title{How AI Assistance Affects Human Skill Development: A Study of Learning with Logic Puzzles}

\author{Shang Wu}
\orcid{1234-5678-9012}
\orcid{0009-0001-4361-8703}
\email{shangw13@uci.edu}
\affiliation{%
  \institution{University of California, Irvine}
  \country{USA}
}

\author{Catarina G Belem}
\orcid{0000-0001-5781-3131}
\email{cbelem@uci.edu}
\affiliation{%
  \institution{University of California, Irvine}
  \country{USA}
}

\author{Shuyuan Fu}
\orcid{0009-0009-8733-3484}
\email{shuyuanf@uci.edu}
\affiliation{%
  \institution{University of California, Irvine}
  \country{USA}
}

\author{Mark Steyvers}
\orcid{0000-0003-1466-5647}
\email{msteyver@uci.edu}
\affiliation{%
  \institution{University of California, Irvine}
  \country{USA}
}

\author{Padhraic Smyth}
\orcid{0000-0001-9971-8378}
\email{smyth@uci.edu}
\affiliation{%
  \institution{University of California, Irvine}
  \country{USA}
}

\renewcommand{\shortauthors}{Wu et al.}

\begin{abstract}
While AI assistance can improve human task performance in the short term, it may also undermine the development of skills in the longer term. We examine this tension in a controlled logic-puzzle experiment involving on-demand AI assistance, where participants complete tasks before, during, and after AI is available. By experimentally varying AI request costs, we find that lower-cost assistance induces more frequent AI use. We also find that participants who request AI assistance during the AI-access phase perform worse at the task after assistance is removed, and their subsequent unassisted performance is overestimated when predicted from earlier AI-assisted performance. We use a Bayesian latent ability model to separate initial ability, post-AI ability, and participant-specific skill change, while estimating how independent reasoning during the AI-access phase relates to skill development. The results show that greater independent problem-solving effort is associated with larger gains in latent ability, consistent with the interpretation that skill development is weaker when AI assistance substitutes for independent reasoning.

\end{abstract}




\maketitle

\section{Introduction}
As AI systems become increasingly used as aids in everyday cognitive activities, an important question is not only whether they improve human cognitive performance but also whether they promote or hinder skill development. While AI assistance can improve immediate task performance in the short term \citep{gajos2022people,karny2024learning}, it may also alter the degree to which people engage with a cognitive task. In particular, one significant concern is that readily available AI assistance could reduce the need for sustained independent reasoning and encourage humans to offload cognitive effort to the AI system \citep{buccinca2021trust,risko2016cognitive,lee2025impact}. This creates a central tension in AI-assisted cognition: the same assistance that helps users solve problems in the short run may weaken the development of the underlying skills needed to solve similar problems independently in the future.

This tension is especially important in settings where learning depends on active engagement. Skill development requires repeated opportunities to reason, test strategies, revise mistakes, and internalize problem structure \citep{anderson1995cognitive}. If AI assistance provides a substitute for these processes, humans may appear to be successful at a task while the assistance is available, but have weaker performance once the assistance is removed \citep{wu2026impact,liu2026ai}. At the same time, AI use is not necessarily harmful in itself. Humans may request assistance while still engaging deeply with the task, and the consequences of AI use may depend on whether it complements or replaces independent problem-solving effort \citep{shen2026ai,lehmann2024ai, poulidis2025action}. Understanding the effect of AI on learning, therefore, requires distinguishing between (a) 
the frequency of AI use, and (b) how users use it.

\begin{figure}[tb] 
  \begin{center}
    \includegraphics[width=0.45\textwidth]{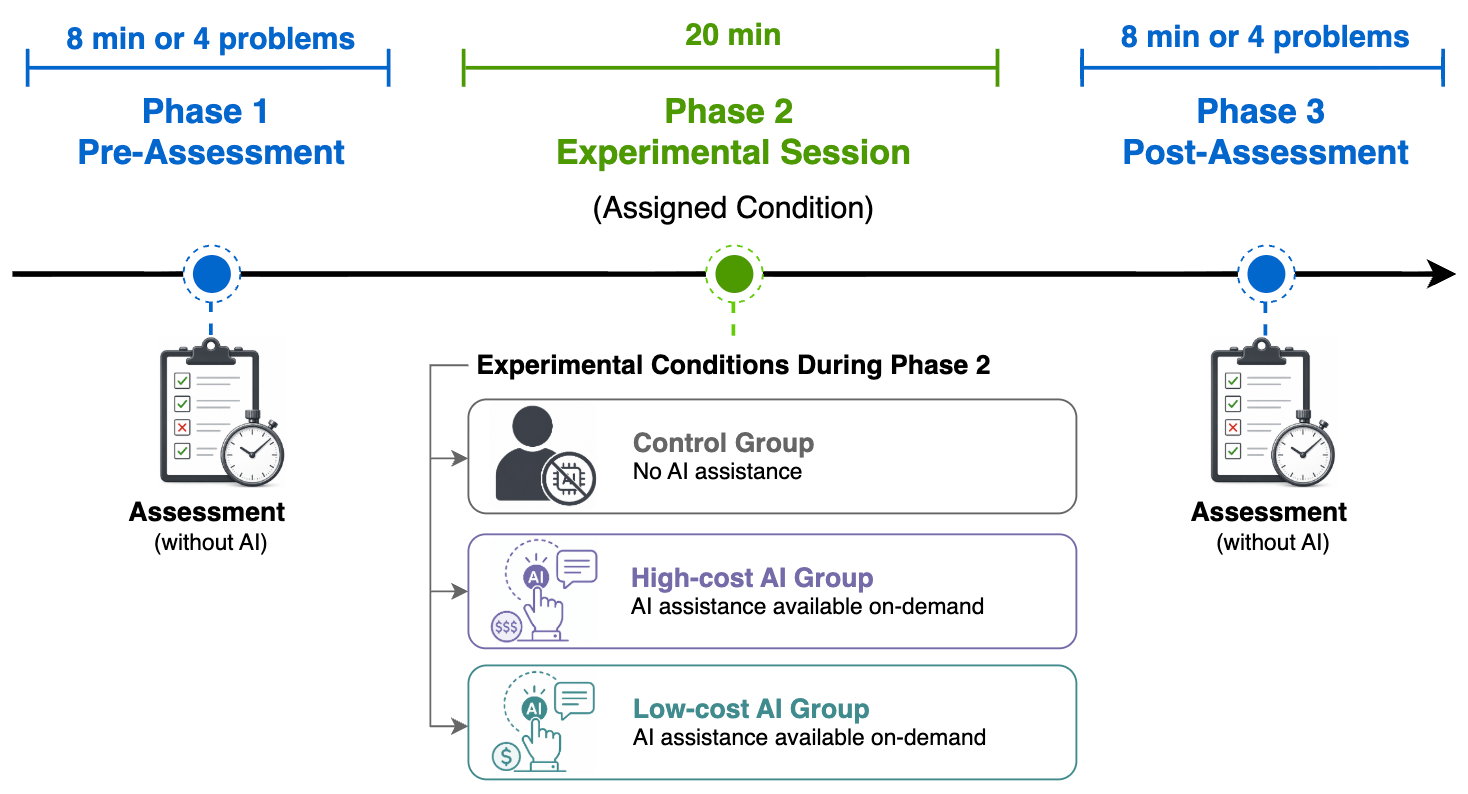}
  \end{center}
\caption{Overview of the three-phase study design. Participants completed Phase~1 and Phase~3 without simulated AI assistance, providing pre- and post-AI measures of unassisted performance. During Phase~2, participants were assigned either to a no-AI control condition or to an AI-access condition with experimentally varied AI request costs.}
\Description{A horizontal timeline shows the three-phase study progressing from a pre-assessment to an experimental phase and then a post-assessment. During the middle phase, participants are randomly assigned to one of three conditions: no AI assistance, high-cost AI assistance, or low-cost AI assistance.}
\label{fig:exp_diagram_flow}
\end{figure}

We study this issue through a controlled logic-puzzle experiment that tracks skill development across three phases: Phase 1, an initial AI-free assessment; Phase 2, an intermediate phase in which AI assistance may be available; and Phase 3, a final AI-free assessment (see Figure \ref{fig:exp_diagram_flow}). During Phase~2, participants can request AI assistance on demand. We experimentally vary the costs of AI requests across conditions to induce different levels of AI engagement across participants.
By comparing participants' performance in Phases~1 and 3, the design allows us to examine how the cost of AI shapes reliance behavior and how participants’ engagement with AI during Phase~2 relates to subsequent performance after assistance is removed.

We investigate relationships between various factors related to individual behavior and performance by fitting a Bayesian latent ability model that treats pre- and post-AI accuracy and response time as indicators of underlying ability, and that models changes in skill levels as a function of observed behavior, performance measures, and initial ability. Our findings suggest that weaker short-term skill development is associated with AI assistance that displaces independent reasoning. Lower-cost access leads participants to use AI more frequently, and participants who request AI perform worse once assistance is removed. The Bayesian model further suggests that short-term skill development is associated less with how often participants request AI help and more with whether they continue to engage in independent reasoning. Participants who devote more effort to independent problem solving show greater gains in latent ability, whereas the frequency of requests for AI assistance is not associated with gains in skill development after accounting for initial ability and independent effort.

This paper contributes to research on human-AI interaction by adding nuance to how AI assistance affects learning. Our findings suggest that, in a controlled logical puzzle task, the central issue is not AI use itself, but whether assistance displaces the independent reasoning through which skills are built. This distinction is particularly important in educational and training settings, where AI support should ideally improve current task performance without weakening future problem-solving ability.

\section{Related Work}
\label{sec:related_work}
Although a large body of work examines human-AI complementarity in joint decision-making contexts \citep{steyvers2022bayesian,zhang2022you,de2024towards}, recent studies have begun to investigate how AI assistance influences subsequent performance once it is no longer available. A common approach uses pre- and post-assessment designs to compare behavior before AI exposure, during AI-assisted work, and after assistance is removed. In mathematical reasoning and reading-comprehension tasks, \citet{liu2026ai} find that AI assistance improves short-run performance but reduces persistence and impairs later independent performance. In programming, \citet{shen2026ai} show that AI assistance can weaken conceptual understanding, code reading, and debugging skills, while also documenting cognitively engaged interaction patterns that preserve learning. Related evidence from writing tasks suggests that LLM assistance may reduce cognitive engagement and weaken later unaided performance \citep{kosmyna2025your}. These findings are consistent with broader work on AI as a ``tool for thought'', which emphasizes that AI systems can either scaffold reasoning or automate cognitive steps that users would otherwise perform themselves \citep{tankelevitch2025understanding}. Together, these studies suggest that AI-assisted success does not necessarily translate into durable skill acquisition, and that AI’s effects on human learning are context-dependent. Our work adds to this growing literature by focusing on whether learners continue to spend time on independent problem-solving during periods when AI assistance is available, and how this independent effort relates to subsequent skill development.

One mechanism linking AI assistance to skill development is cognitive offloading: the use of external tools to reduce internal cognitive effort \citep{risko2016cognitive}. Offloading can improve immediate performance, but it may also reduce learning when externally supported operations replace the reasoning, memory, or problem-solving processes through which skill is acquired \citep{grinschgl2021consequences,sparrow2011google}. This connects directly to work on AI reliance, which studies when users request, accept, reject, or defer to AI assistance \citep{madhavan2007effects,hoff2015trust}. Prior work shows that reliance depends on features of the task environment, such as AI accuracy, timing, and time pressure \citep{cao2023time,nourani2020role,swaroop2024accuracy}. Reliance, however, is not necessarily harmful: on-demand assistance can act as a cognitive forcing mechanism that reduces over-reliance \citep{buccinca2021trust}, and learning-oriented assistance can be beneficial when it scaffolds understanding rather than allowing users to bypass the task \citep{gajos2022people,baker2004off}. Thus, a key distinction is not simply whether users rely on AI, but whether AI use preserves or displaces the cognitive work needed for learning.

Prior work also suggests that the consequences of AI assistance depend on how the system changes the user's role in the task. In chess, \citet{poulidis2025action} show that the effects of AI assistance depend on whether it prescribes actions or instead directs users’ attention, reinforcing the distinction between AI that substitutes for vs. complements human reasoning. 
In our earlier work \cite{wu2026impact}, we explored the role of AI informativeness and usage in logical reasoning. In this paper we build on this framework to focus more directly on the behavioral mechanisms linking AI assistance to skill development. Specifically, we distinguish between the frequency and timing of AI requests and the extent to which participants preserve independent problem-solving effort when assistance is available. In this context, we develop a Bayesian latent ability model that allows us to disentangle the different factors affecting skill development over time. This allows us to examine whether weaker subsequent learning is better explained by AI use itself or by the displacement of independent reasoning during the AI-access phase.

\section{Experiment}

To understand how AI assistance affects human skill development, we conducted a controlled user study in the specific context of solving logic puzzles under time constraints, where participants could autonomously request AI assistance while solving problems. By experimentally varying the cost of AI requests, the design allows us to examine how different help-seeking environments shape AI engagement, task performance, and learning.

\subsection{Task Description}
\label{subsection:exp_1_task_description}
\begin{figure}[tb]
    \centering

    \begin{subfigure}{\columnwidth}
        \centering
        \includegraphics[width=\linewidth]
        {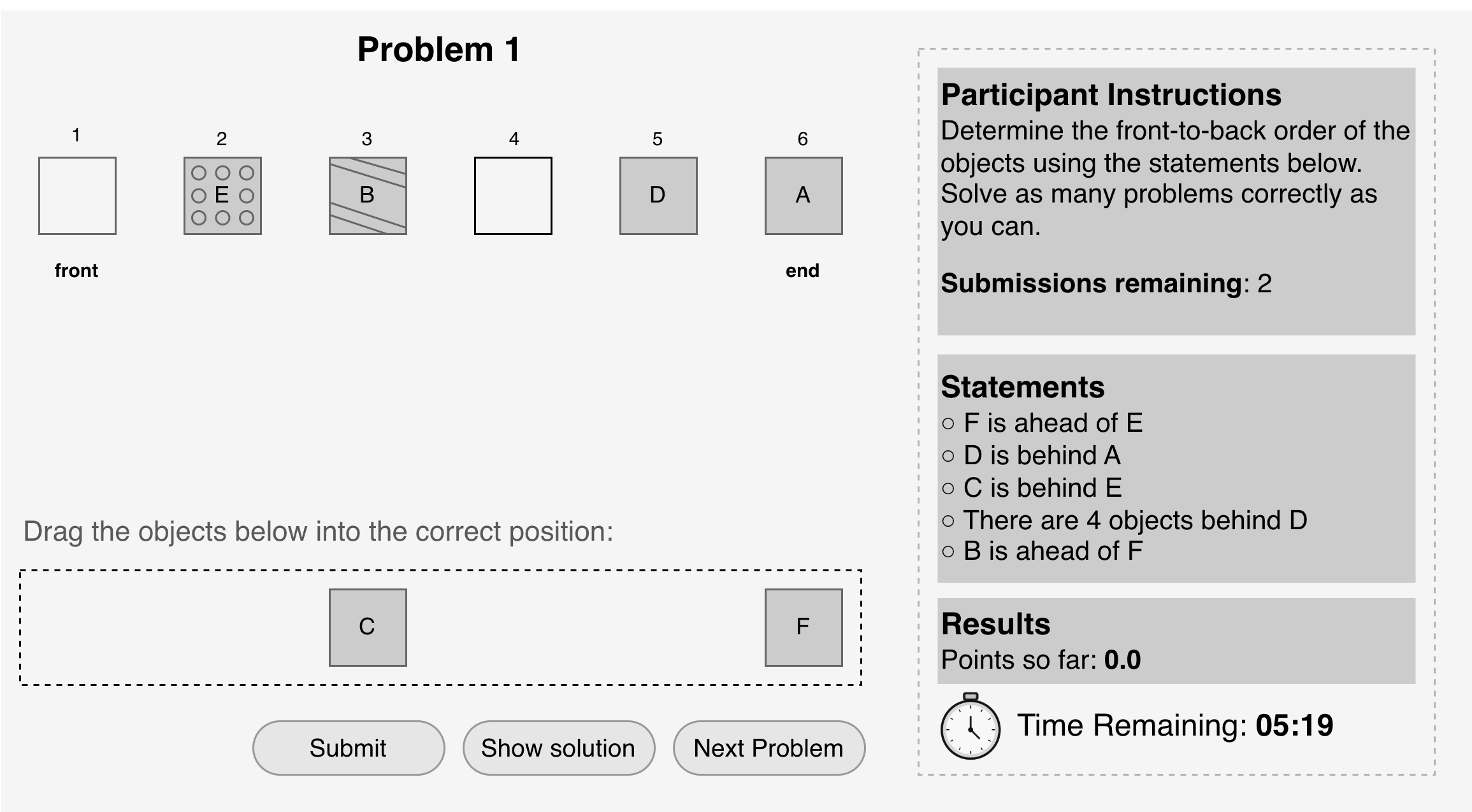}
        \caption{Without AI assistance}
        \label{fig:screenshot_noAI}
    \end{subfigure}

    \vspace{0.6em}

    \begin{subfigure}{\columnwidth}
        \centering
        \includegraphics[width=\linewidth]
        {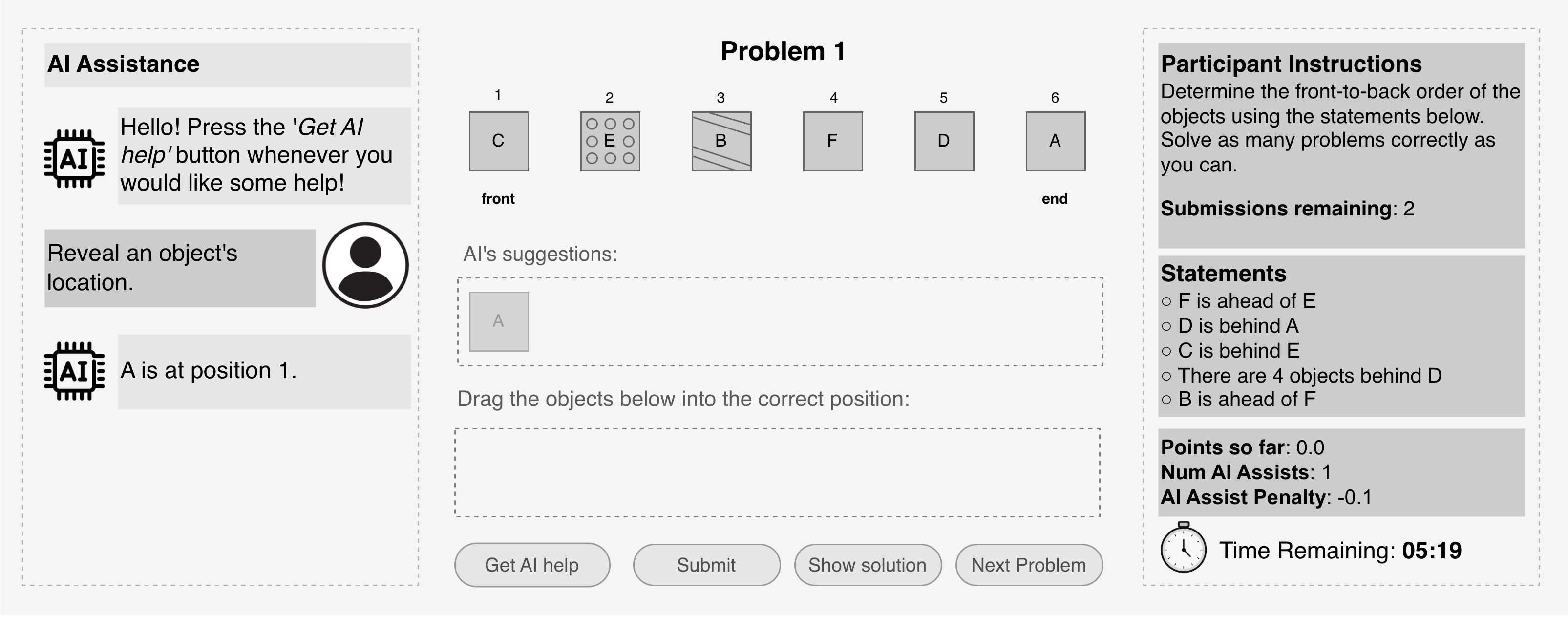}
        \caption{With AI assistance}
        \label{fig:screenshot_withAI}
    \end{subfigure}

    \caption{User-interface examples from the logic-based puzzle task}
    \Description{Examples of two user interfaces for the logic-based puzzle task, consisting of 6 objects. (a) The user interface without AI assistance; (b) The user interface with AI assistance, identical to (a) except for an added dialog on the left that displays the AI suggestion as text and displays the location of a randomly chosen object on the screen. }
    \label{fig:ui_examples}
\end{figure}

We build on a logic-puzzle paradigm used in prior work on AI-assisted skill development \citep{wu2026impact}, but modify the AI assistance format to allow for cost-driven help-seeking. In each problem, participants were presented with six objects and a set of five logical constraints that jointly define a unique ordering (Figure \ref{fig:screenshot_noAI}). Participants solved the problem by arranging the six objects according to these constraints. An answer was deemed correct (from the user's perspective) only if all six objects were placed in the correct positions. Participants could submit up to two attempts per problem. After the first attempt, participants received feedback indicating the number of correctly positioned objects, but not which specific objects were correctly positioned. Given this feedback, participants could either revise and resubmit their response or accept the initial attempt as final and proceed to the next problem. The correct solution was revealed after each problem to facilitate learning across problems.

While more open-ended and more realistic tasks could in principle be used in a study like this, this more controlled setting reduces variability in terms of prior human knowledge and supports a clearer focus on problem-solving behavior under optional AI assistance. The task was designed to be intuitive for participants without specialized training, while still requiring logical reasoning under time constraints, reflecting decision settings in which both speed and accuracy are important. In addition, some objects contained recurring visual markers that were associated with specific positions, providing hidden structural cues that participants could learn to exploit across problems.

The study employed a three-phase design that separates baseline performance, AI-assisted task completion, and post-assistance performance, allowing us to examine how AI request costs relate to AI usage patterns and subsequent skill development (Figure \ref{fig:exp_diagram_flow}). Phase 1 served as a \textit{pre-AI assessment} and lasted 8 minutes or until participants completed at least four problems. Phase 2 lasted 20 minutes and introduced the experimental conditions, including optional AI assistance under different costs as described below in Section~\ref{subsection:exp_1_conditions}. Phase 3 mirrored Phase 1 as a \textit{post-AI assessment} without AI assistance. Problem order was randomized within each phase to mitigate confounding effects of problem difficulty on learning.

Participants earned one point for each correctly solved problem. During Phase 2, points were deducted for each AI request according to the assigned condition. Bonus payments were calculated at \$0.15 per point in Phases 1 and 2 (up to \$2 total) and \$0.18 per point in Phase 3 (up to \$1.5).

\subsection{Conditions}
\label{subsection:exp_1_conditions}
We used a between-subjects design in which participants were randomly assigned to one of three conditions that varied the availability and cost of AI assistance during Phase 2. AI assistance was available only in Phase 2, and participants in AI-available conditions could request it at any time while solving the puzzles. Each request revealed the location of one randomly selected object and incurred a condition-specific point deduction, creating a trade-off between obtaining help and maximizing rewards. Participants were informed of the applicable AI request cost structure in the instructions before starting the study.

Figure \ref{fig:screenshot_withAI} shows the user interface for an AI-enabled condition. This interface is representative of all AI-enabled conditions, which share the same interaction design and differ only in the cost of AI requests. Participants could click ``Get AI help'' to reveal the location of a randomly selected object in the puzzle. Each request updated the right-hand panel by incrementing the number of assists and the deducted points.

Participants were randomly assigned to one of the following conditions:
\begin{itemize}
    \item \textit{\textbf{No-AI}}: No AI assistance was available, and the AI panel was not shown (see Figure \ref{fig:screenshot_noAI}).
    \item \textit{\textbf{Low-cost AI}}: Each AI request incurred a 0.1-point deduction.
    \item \textit{\textbf{High-cost AI}}: Each AI request incurred a 0.18-point deduction.
\end{itemize}

To ensure that variation in AI quality did not confound the effect of AI cost on participants' help-seeking behavior, we fixed the simulated AI agent to be perfectly correct (100\%). Under this design, each AI request consistently returned the correct location of a randomly selected object, allowing us to attribute behavioral differences across AI-access conditions to request cost rather than output quality.
This design choice aligns with prior work in human–AI interaction that uses simulated agents to isolate specific behavioral mechanisms (e.g., \cite{srivastava2022improving, buccinca2021trust, gajos2022people,vasconcelos2023explanations}). Participants were not informed of the AI’s accuracy and were instructed only that AI assistance would be available during Phase 2 at a cost.

\subsection{Procedure}
After providing informed consent, participants were randomly assigned to one of the three conditions and presented with task instructions. To ensure task understanding, participants completed two comprehension checks, and only those who passed both proceeded to the main study.

The main study consisted of three sequential phases, separated by short breaks, during which participants completed a set of logic puzzle problems under their assigned condition. To monitor data quality during task performance, an attention check was embedded in Phase 2.
After completing all phases, participants completed a short post-study survey about their experience, perceived effort, strategy use, prior task familiarity, and, when applicable, their perceptions and timing of AI use (see Supplementary Material~\ref{appendix:post-study_survey} % B.1 
for details).

\subsection{Participants}
We recruited 150 English-speaking adults (18+) via Prolific, restricted to adults residing in the U.S. with at least an undergraduate degree. We excluded 26 participants: 1 failed the attention check, 14 showed inattentiveness (off-screen for $\geq 5$ seconds on over three problems), 8 had inconsistent self-reported and logged AI use, 2 were extreme time outliers, and 1 showed low-effort responses with near-zero accuracy.\footnote{Off-screen time is total problem time minus active on-screen time. Inconsistent AI-use reporting refers to disagreement between post-study reports and usage logs. The low-effort participant expressed dislike for the study and answered randomly throughout.} The final sample consisted of 124 participants (42 \textit{No-AI}, 43 \textit{Low-cost AI}, 39 \textit{High-cost AI}).

Participants received \$9 for participation (median completion time: 56 minutes) and could earn up to an additional \$3.50 in performance-based bonuses. The study was approved by the Institutional Review Board (IRB) at the University of California, Irvine (protocol \#7206).

The final sample consisted of 124 participants with a mean age of 40 years (SD = 12). The sample included 59 male participants, 64 female participants, and 1 participant who preferred not to disclose gender. In terms of education, 66 participants held an undergraduate degree and 58 held a master's degree or higher.

\subsection{Measures}
\label{subsection:exp1_measure}

We study both observed task performance and latent skill development. For observed performance, we use \textbf{accuracy}, defined as the number of objects (0--6) placed in the correct position for a problem, and \textbf{response time}, measured in seconds. Both are computed per participant for each of the 3 phases, averaging over the different problems that a participant attempted during a phase. To summarize the speed-accuracy tradeoff, we also compute \textbf{reward rate}, as used in cognitive modeling \citep{standage2015toward}. For each participant in each phase, reward rate is defined as the average accuracy divided by the average response time, in units of correct objects per minute, where higher values indicate better performance.  

To characterize reliance on AI during Phase~2, we measure \textbf{AI usage} as the total number of assistance requests made by a participant.
We also measure independent problem-solving effort using \textbf{solo share}. At the problem level, this measure is defined as the fraction of response time spent working independently before requesting AI assistance. If no AI assistance is requested, the time until final submission is counted as the time working independently. Participant-level solo share is then defined as the average problem-level solo share across Phase~2 problems.

In Section~\ref{sec:empirical_results} we use reward rate, accuracy, and response time to empirically summarize performance, to characterize AI reliance behavior, and to examine whether Phase~2 performance predicts subsequent unassisted performance differently across levels of AI use. Our Bayesian analyses in Section~\ref{sec:bayesian_analysis} use Phase~1 and Phase~3 performance as observed signals of latent ability, and use Phase~2 solo share as the main behavioral measure to predict the final (Phase~3) skill levels of participants, both with and without AI assistance.

\section{User Study Results}
We begin below with a high-level summary of results from our user study, followed by a more detailed Bayesian analysis to quantify the effect of AI assistance on skill development at both individual and population-wide levels. 

\subsection{Empirical Findings}
\label{sec:empirical_results}

\begin{figure}[tb]
    \centering
    \includegraphics[
        width=0.65\columnwidth,
        keepaspectratio
    ]{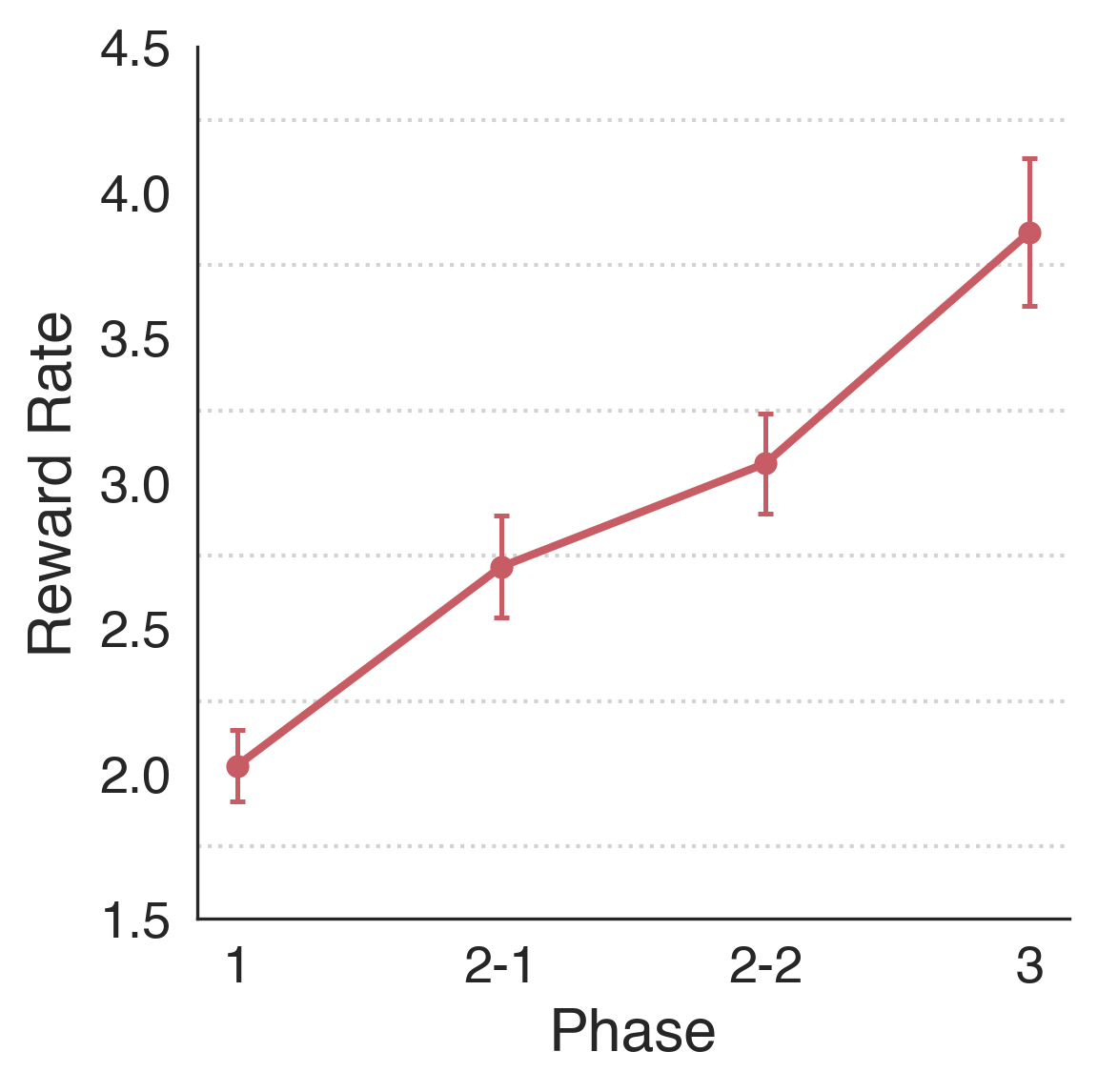}
    \caption{Reward rate across phases for participants with zero AI use in Phase~2. Labels 2-1 and 2-2 denote the first and second halves of Phase~2. Error bars indicate standard errors.}
    \Description{A line plot shows reward rate increasing steadily across the four displayed time points, from 2.03 in Phase 1 to 3.86 in Phase 3, with intermediate increases during the two halves of Phase 2.}
    \label{fig:exp1_general_learning}
\end{figure}

\paragraph{Skill Development:} To examine whether participants in general showed task-specific skill improvement over time, we show in Figure~\ref{fig:exp1_general_learning} the average reward rate across phases for participants who did not request any AI assistance. The unassisted task performance of this set of participants improved substantially 
from the pre-AI assessment (Phase~1) to the post-AI assessment (Phase~3). The mean reward rate increased from 2.03 correct objects per minute in Phase~1 to 3.86 in Phase~3, corresponding to a 90.2\%   increase in the number of correct objects per minute ($N=75$, $t=7.40$, $p<0.01$), illustrating substantial short-term improvement in participants' skills on this task.

\begin{figure}[tb]
    \centering
    \includegraphics[
        width=0.65\columnwidth,
        keepaspectratio
    ]{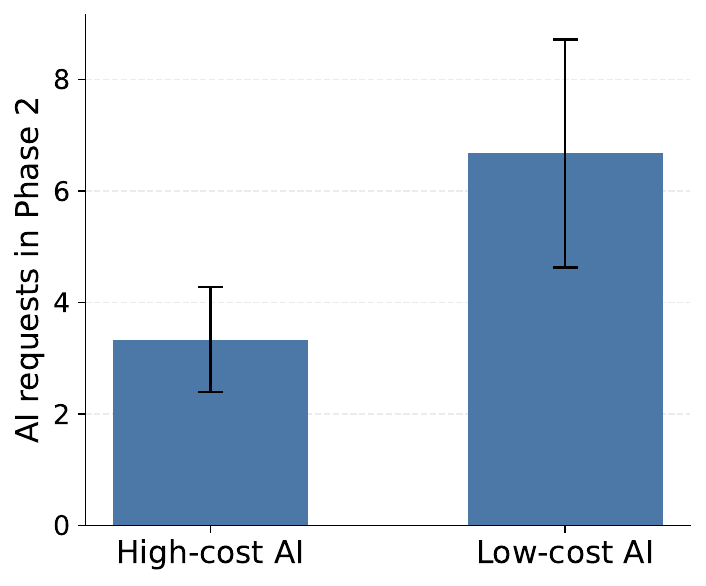}
    \caption{Average AI usage in Phase~2 by AI cost condition. Error bars indicate standard errors.}
    \Description{A two-bar chart shows that participants in the low-cost AI condition made roughly twice as many AI requests as participants in the high-cost AI condition,  6.67 vs. 3.33 requests on average.}
    \label{fig:exp1_ai_usage_mean}
\end{figure}

\paragraph{Effects of AI-cost condition:}
We next examine the effects of the randomized AI-cost manipulation. Phase~1 reward rates were well balanced across conditions (ANOVA: $F=0.20$, $p=0.82$). As shown in Figure~\ref{fig:exp1_ai_usage_mean}, participants in the \textit{Low-cost AI} condition made significantly more AI requests during Phase~2 than those in the \textit{High-cost AI} condition (6.67 vs.\ 3.33 requests; one-sided Welch test: $t=1.48$, $p<0.10$). From Phase~1 to Phase~3, reward rate increased by 1.95 in the \textit{No-AI} condition, 1.66 in the \textit{High-cost AI} condition, and 1.32 in the \textit{Low-cost AI} condition; the increase was smaller in the \textit{Low-cost AI} than in the \textit{No-AI} condition in a one-sided $t$-test ($t=1.30$, $p<0.10$). In Phase~3, accuracy did not differ across conditions (ANOVA: $F=0.097$, $p=0.91$), while response times were longer in the \textit{Low-cost AI} condition ($M=111.81$s, $SE=6.13$) than in the \textit{High-cost AI} ($M=86.78$s, $SE=4.18$; $p<0.01$) or \textit{No-AI} condition ($M=92.21$s, $SE=4.88$; $p=0.01$).

\begin{figure}[tb]
    \centering
    \includegraphics[
        width=0.65\columnwidth,
        keepaspectratio
    ]{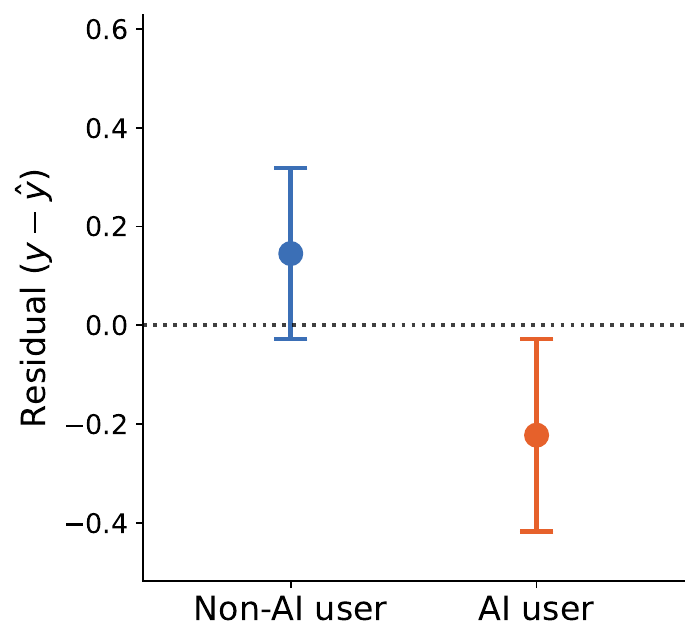}
    \caption{Residuals by AI-use status. Residuals are defined as observed minus predicted Phase~3 reward rate, $y - \hat{y}$; error bars indicate standard errors.}
    \Description{An error-bar plot compares prediction residuals for non-AI users and AI users. The average residual is 0.15 for non-AI users and -0.22 for AI users, indicating opposite directions of prediction error across the two groups.}
    \label{fig:phase3_residual_ai_use}
\end{figure}

\paragraph{Post-AI Performance:} 
Descriptively, participants who requested AI assistance in Phase~2 had a lower average reward rate (3.42 correct items/minute) in Phase~3 (where no AI assistance was available) compared to the average reward rate in Phase~3 (3.86 correct items/minute) of participants who did not request AI in Phase~2. To explore this further, we use a regression model to predict Phase~3 reward rate as a function of Phase~2 reward rate and find that predictions are systematically biased by AI use. Specifically, as shown in Figure \ref{fig:phase3_residual_ai_use}, for participants who did not request AI, Phase~3 performance was underestimated by 0.15 reward-rate units on average. In contrast, for AI users, Phase~3 performance was overestimated by 0.22 reward-rate units on average. In other words, the apparent performance of participants while using AI (in Phase~2) systematically overestimates their subsequent performance when conducting the task on their own without AI assistance (in Phase~3).

These patterns suggest that Phase~2 AI-use behavior is associated with differences in short-term skill improvement from the pre-AI phase (Phase~1) to the post-AI phase (Phase~3). We investigate these relationships further below via a Bayesian modeling approach, separating initial ability, post-AI ability, and latent skill change while incorporating Phase~2 AI usage behavior.

\subsection{Bayesian Analysis}
\label{sec:bayesian_analysis}

We use a Bayesian model to infer latent individual ability and skill development, conditioned on observed individual-level data (accuracy, response time, AI usage, solo share) across the three experiment phases. 
This approach follows a long tradition in education and psychometrics in modeling ability as a latent trait inferred from observed performance \citep{lord2008statistical,embretson2025item}. 
It is useful in our setting because observed performance is a noisy proxy for ability: accuracy and response time provide complementary signals, and the same observed Phase~3 performance can correspond to different amounts of learning depending on participants' Phase~1 baseline performance and Phase~2 AI reliance behavior. A Bayesian latent ability framework allows us to separate initial ability from post-AI ability and participant-level skill change, while propagating uncertainty in both individual- and population-level quantities.
In particular, we use the model to infer each participant's individual latent initial ability, latent post-AI ability, and latent skill change, as well as to estimate how short-term skill change is associated with independent problem-solving effort in Phase~2. We outline below the structure of the model in correspondence with the three phases of the experiment. We standardize all observed performance measures prior to fitting the model to improve the interpretability of coefficients across variables measured on different scales \cite{gelman2013bayesian, gelman2007data}, and we use weakly informative priors as outlined in the Supplementary Material~\ref{app:prior_specification}.
%A.1. 

\subsubsection{A Bayesian latent ability model of short-term skill development, with and without AI assistance}
\label{subsec:bayesian_model}
Beginning with Phase~1 (pre-AI, initial assessment), 
for individual $i$, let $\theta_{i1}$ denote latent initial ability, with a prior $\theta_{i1} \sim \mathcal{N}(0,1)$. 
We model an individual's Phase~1 outcomes (accuracy $A_{i1}$ and log response time  $R_{i1}$) as   conditionally independent noisy functions of their latent ability $\theta_{i1}$: 
\begin{equation}
\begin{aligned}
A_{i1} \mid \theta_{i1}, \mu_A, \sigma_A \sim \mathcal{N}(\mu_A + \theta_{i1}, \sigma_A) 
\\
R_{i1} \mid \theta_{i1}, \mu_R, \sigma_R \sim \mathcal{N}(\mu_R - \theta_{i1}, \sigma_R)
\end{aligned}
\end{equation}

where the effect of $\theta_{i1}$ is reversed for the two variables (in terms of the signs in the mean terms) to reflect the expectation that higher values of ability will correspond (on average) to higher accuracy and to lower response times. 
Here, $\mu_A$ and $\mu_R$ are measurement intercepts for accuracy and log response time, respectively, and $\sigma_A$ and $\sigma_R$ capture residual measurement variation. Because $A_{i1}$ and $R_{i1}$ are phase-level measures aggregated over the problems completed by each participant, the model does not separately estimate problem-level effects. Problems were designed to be comparable in difficulty and their order was randomized within each phase, so any remaining variation in problem difficulty, order, or the number of problems completed is treated as part of the residual measurement variation. These population-level parameters are inferred jointly during Bayesian inference along with the individual-level latent abilities and skill changes introduced below.

For Phase~2, where AI is available to some participants, we use solo share $\mathrm{Solo}_i$ as the observed behavioral measure in our main model, where $\mathrm{Solo}_i$ is the participant-level fraction of Phase~2 problem-solving time that participant $i$ spends reasoning independently prior to (or without) AI assistance (see Section \ref{subsection:exp1_measure}).  We model participant-specific latent skill change, $\delta_i$, as a function of initial ability and the Phase~2 engagement measure $\mathrm{Solo}_i$:
\begin{equation}
\delta_i \mid \theta_{i1}, \mathrm{Solo}_i
\sim
\mathcal{N}\left(
\alpha_0
+ \alpha_{\theta}\theta_{i1}
+ \alpha_{\text{solo}}\mathrm{Solo}_{i},
\ \sigma_{\delta}
\right).
\label{eq:latent_gain_main}
\end{equation} 
Since $\delta_i$ is not constrained to be positive, the model can accommodate either skill improvement or skill decline after AI assistance. The intercept $\alpha_0$ represents the expected latent skill change for a participant with average initial ability and average solo share. The coefficient $\alpha_{\theta}$ allows latent skill change to vary with initial ability, accounting for the possibility that lower- and higher-ability participants have different room to improve or learn at different rates (e.g., allowing for saturation effects for individuals with very high initial ability). The $\alpha_{\mathrm{solo}}$ coefficient captures the association of independent reasoning with latent skill change, conditioned on initial ability. Note that because solo share reflects observed participant behavior rather than a randomized treatment, this coefficient should be interpreted as associational rather than causal. The parameter $\sigma_{\delta}$ reflects residual heterogeneous variability in latent skill change that is not explained by initial ability or solo share. 
An alternative option would be to model latent skill change using AI usage as a Phase~2 behavioral predictor, either in addition to or instead of solo share. We use solo share in the main model because, as shown later, it led to better held-out predictive performance than AI request frequency.

For Phase~3, where AI is not available (as in Phase~1), let $A_{i3}$ denote Phase~3 accuracy and $R_{i3}$ denote Phase~3 log response time. Phase~3 outcomes are modeled analogously to Phase~1, i.e., as noisy functions of an individual's latent Phase~3 ability $\theta_{i3}$:
\begin{equation}
\begin{aligned}
A_{i3} \mid \theta_{i3}, \mu_A, \sigma_A\sim \mathcal{N}(\mu_A + \theta_{i3}, \sigma_A) \\
R_{i3} \mid \theta_{i3}, \mu_R, \sigma_R \sim \mathcal{N}(\mu_R - \theta_{i3}, \sigma_R)
\end{aligned}
\end{equation}
We model an individual's latent ability $\theta_{i3}$ as the deterministic sum of their initial ability $\theta_{i1}$ offset by the individual's change in ability $\delta_i$, i.e., $\theta_{i3} = \theta_{i1} + \delta_i$, where posterior uncertainty about $\theta_{i3}$ is driven by uncertainty about $\theta_{i1}$ and $\delta_i$.

\begin{figure}
\centering
    \includegraphics[width=0.95\linewidth]{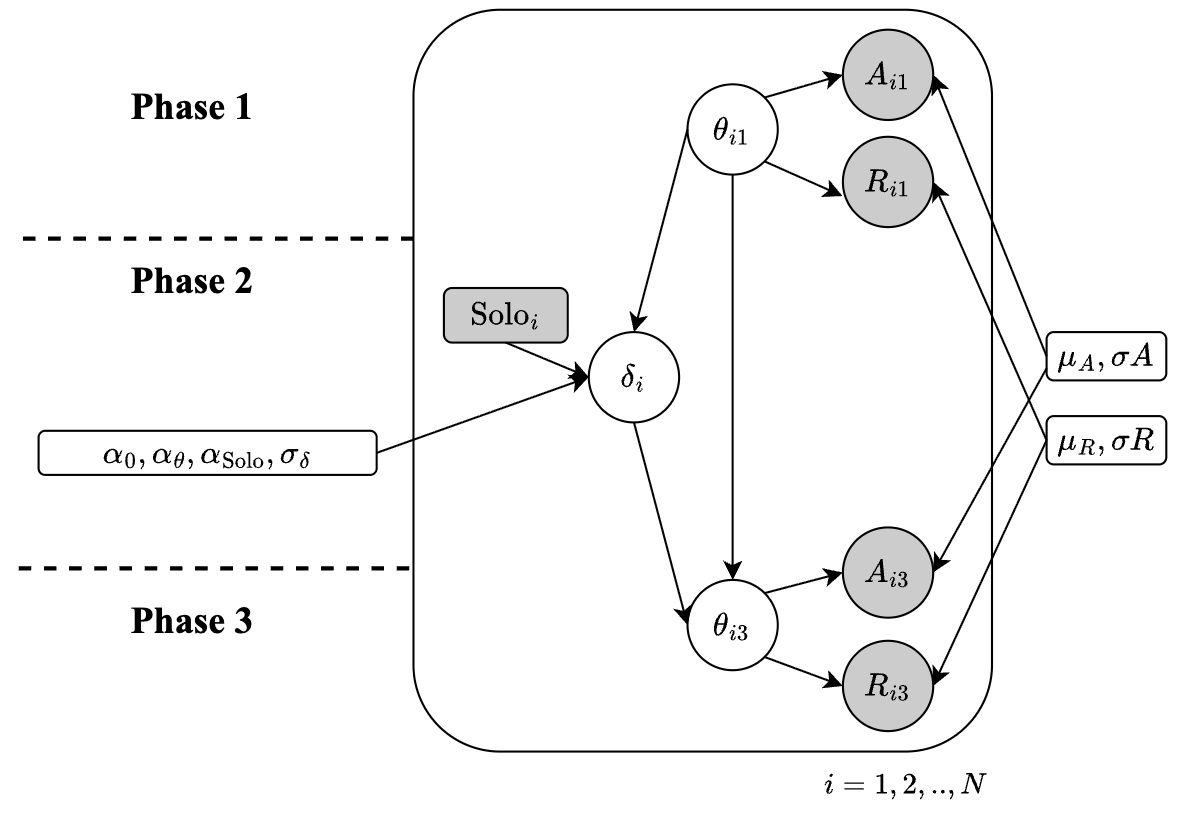}
    \caption{Bayesian model of latent skill development. Phase~1 and Phase~3 accuracy and response time are modeled as noisy measurements of latent abilities $\theta_{i1}$ and $\theta_{i3}$. Participant-specific skill change $\delta_i$ depends on initial ability and solo share. Shaded nodes are observed variables, unshaded nodes are latent variables or unknown parameters, and the plate denotes participants $i=1,\ldots,N$.}
    
    \Description{A plate diagram shows the Bayesian model linking initial latent ability to Phase~1 accuracy and response time, and linking Phase~3 latent ability to the corresponding Phase~3 measures. Initial ability and Phase~2 solo share jointly predict participant-specific skill change, which in turn connects initial ability to post-AI ability.}
    \label{fig:latent_gain_graphical_model}

\end{figure}

We summarize the overall structure of the Bayesian model as a directed graphical model in Figure~\ref{fig:latent_gain_graphical_model}, where shaded nodes denote observed variables and unshaded nodes denote latent variables and unknown parameters in the model. The nodes inside the plate reflect replication over the $N$ participants, where nodes inside the plate are at the individual level (with subscript $i$) and are independent (across the $N$ individuals) conditioned on the population-level parameters outside the plate. The accuracy $A$ and response time $R$ in Phases 1 and 3 depend on the corresponding latent abilities $\theta_{i1}$ and $\theta_{i3}$ for these Phases. The dependence of $\theta_{i3}$ on $\delta_i$ and $\theta_{i1}$ represents the latent transition from initial ability to post-AI ability. 
Finally, latent ability change, $\delta_i$, is modeled as a function of initial ability $\theta_{i1}$ and solo share, allowing the model to capture associations between individual skill development and the amount of independent reasoning participants engage in during Phase~2. 
The graphical model represents the probabilistic dependencies assumed in the model and is not intended as 
a causal model of how Phase~2 behavior affects skill change.
We note that the model does not include variables for observed participant Phase~2 accuracy $A_{i2}$ or response time $R_{i2}$. We omit these measures to keep the specification focused on whether Phase~2 independent reasoning is associated with subsequent latent skill change beyond initial ability, rather than modeling Phase~2 task performance directly.

Given the model described above, Bayesian inference proceeds by conditioning on
$\mathcal{D}=\{A_{i1},\allowbreak A_{i3},\allowbreak R_{i1},\allowbreak R_{i3},\allowbreak \mathrm{Solo}_i\}_{i=1}^{N}$,
and estimating the joint posterior distribution over all unknown quantities. Because the likelihood and priors do not yield a conjugate closed-form posterior, we approximate the posterior using Markov Chain Monte Carlo (MCMC) (see implementation details in Supplementary Material~\ref{app:mcmc_inference}). %A.2). 
Standard MCMC diagnostics indicated satisfactory convergence, and we additionally assess model fit using posterior predictive checks and the uncertainty of held-out predictive comparisons; details are reported in Supplementary Material~\ref{app:model_checks}.
%A.3. 

We note that the randomized AI-cost condition (low-cost, high-cost) affects participants' opportunities and incentives to request assistance, and this condition could be modeled as an upstream parent of solo share in the model. Because the primary specification conditions directly on this observed behavioral measure, we omit the condition node from the graphical model in Figure~\ref{fig:latent_gain_graphical_model} for simplicity. In Supplementary Material~\ref{app:condition_controls} %A.4 
and Supplementary Material~\ref{app:usage_only_model}, %A.5, 
we report the results of alternative model specifications that add AI-cost condition controls and that replace solo share with AI usage, and find that these alternative model specifications do not add any additional explanatory power to the model.

\begin{figure}[t]
    \centering
    \includegraphics[width=0.85\linewidth]{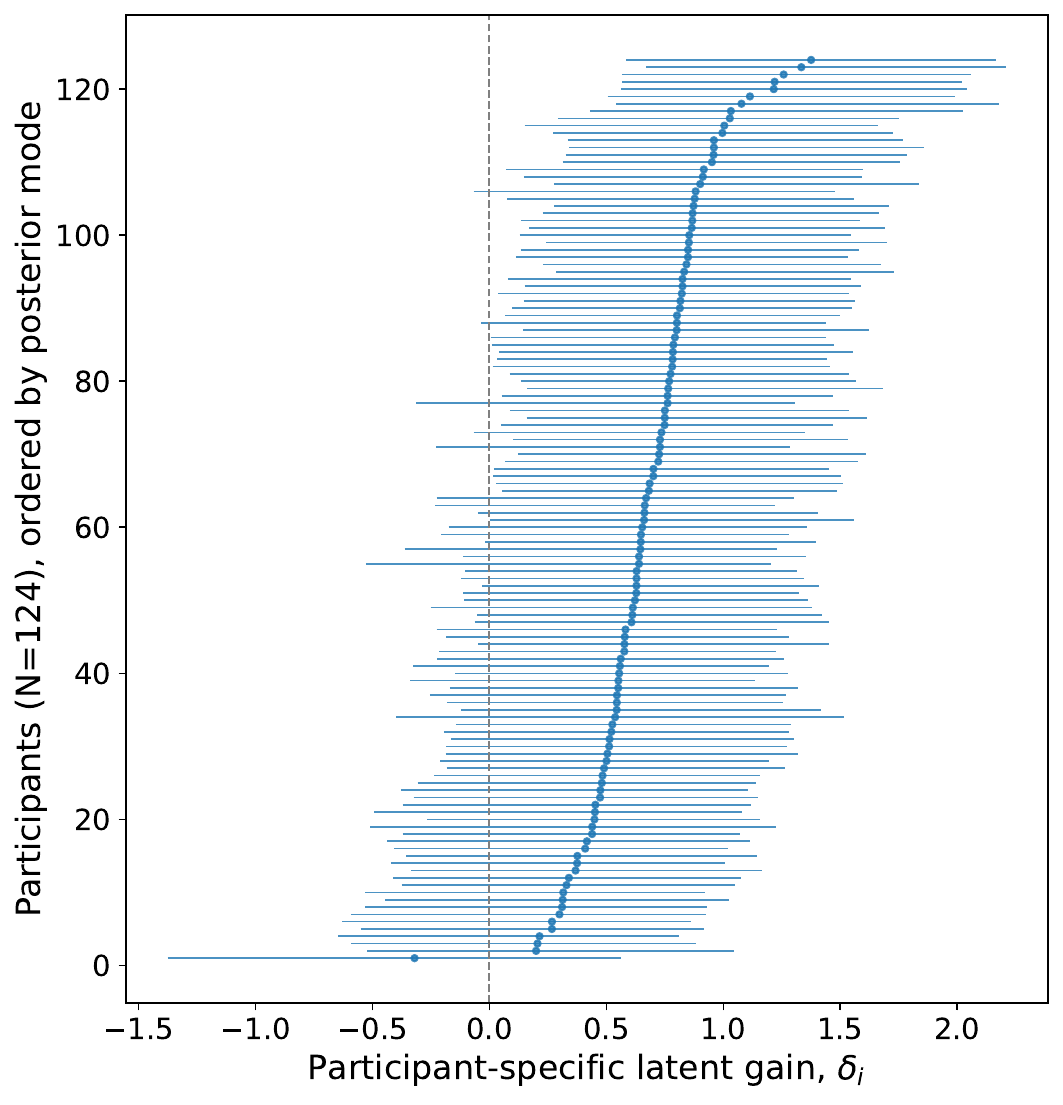}
    \caption{Posterior modes and 95\% credible intervals of participant-specific latent ability changes, $\delta_i$, ordered by posterior mode. The vertical dashed line indicates zero change.}
    \Description{Participant-specific posterior estimates of latent skill change are ordered from smallest to largest. Most posterior modes lie to the right of the zero-change line, while many credible intervals are wide, showing substantial uncertainty and heterogeneity across participants.}
    \label{fig:participant_delta_modes}
\end{figure}

\subsubsection{Independent reasoning is more predictive of short-term skill development than AI request frequency}
\label{subsec:bayesian_results}
We summarize the results obtained from the posterior samples after running MCMC inference for the model described in Figure~\ref{fig:latent_gain_graphical_model}. Figure~\ref{fig:participant_delta_modes} visualizes the posterior modes and 95\% credible intervals of participant-specific latent ability changes, $\delta_i$. Almost all participant-level posterior modes are positive, indicating positive estimated skill development for most participants. This is consistent with the earlier empirical observations of skill development at the population level in Figure~\ref{fig:exp1_general_learning}. The credible intervals are wide, indicating substantial uncertainty and heterogeneity in individual-level skill development, but nonetheless broadly indicate positive skill gains.

\begin{figure}[htbp]
    \centering

    \begin{subfigure}{0.4\textwidth}
        \centering
        \includegraphics[width=0.95\linewidth]{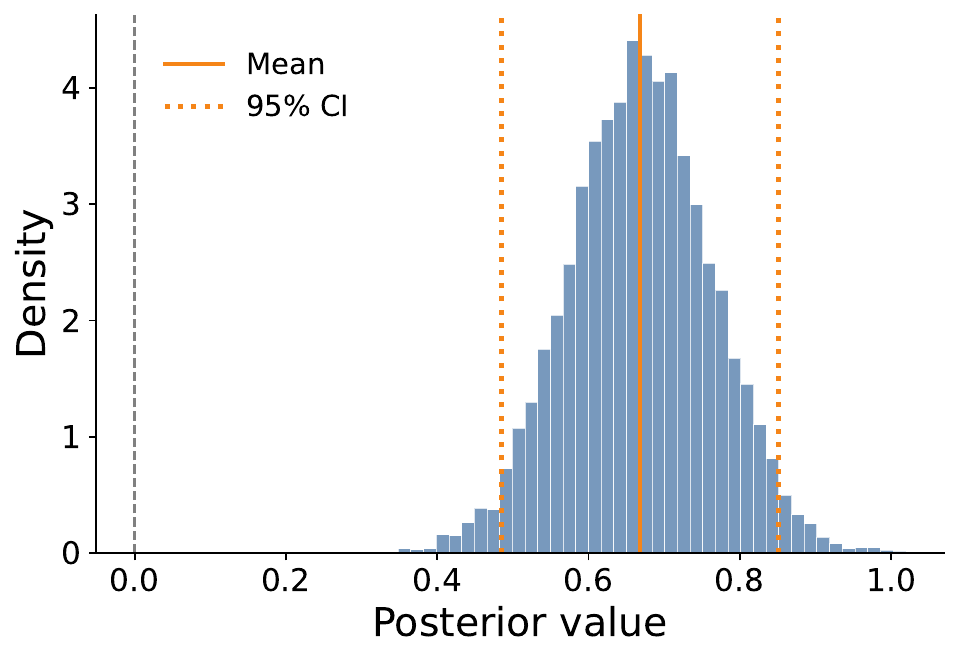}
        \caption{$\alpha_0$: Average expected skill change}
        \label{fig:alpha0_hist}
    \end{subfigure}
    \hfill
    \begin{subfigure}{0.4\textwidth}
        \centering
        \includegraphics[width=0.95\linewidth]{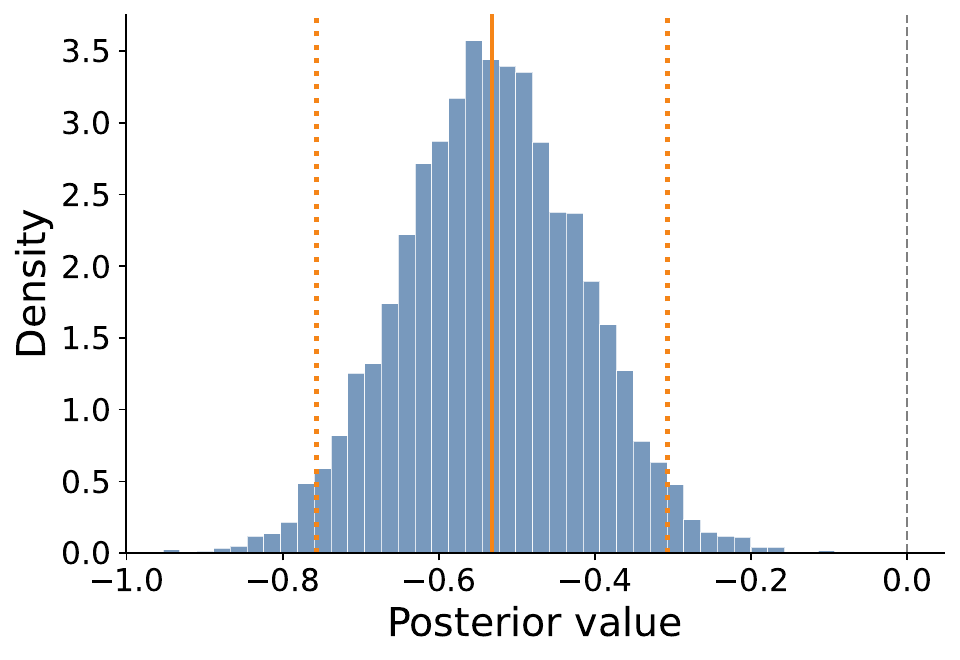}
        \caption{$\alpha_{\theta}$: Impact of initial ability on skill change}
        \label{fig:alpha_theta_hist}
    \end{subfigure}
    \hfill
    \begin{subfigure}{0.4\textwidth}
        \centering
        \includegraphics[width=0.95\linewidth]{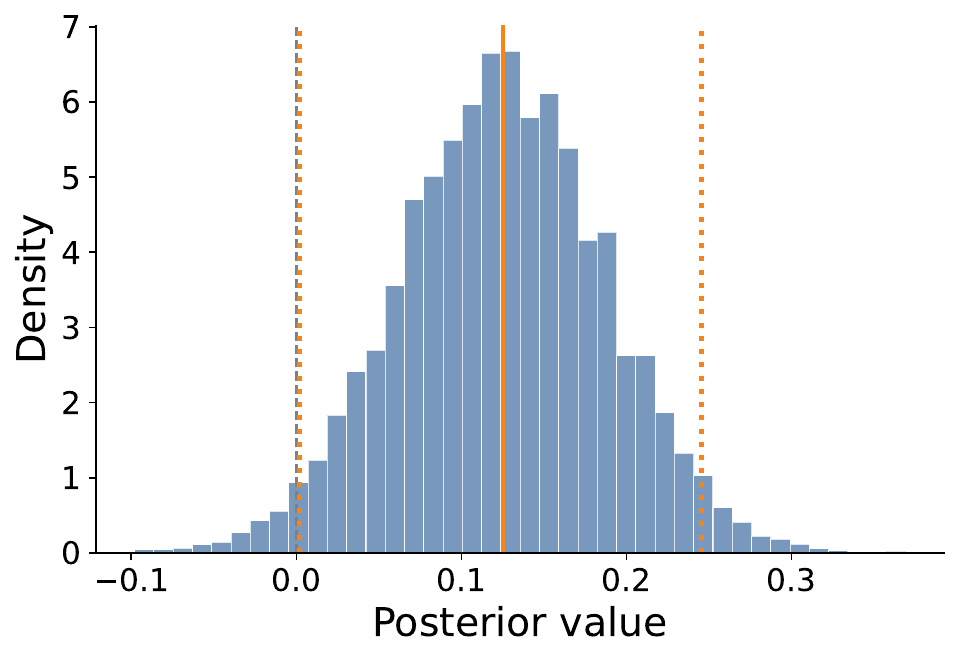}
        \caption{$\alpha_{\mathrm{solo}}$: Solo share}
        \label{fig:alpha_solo_hist}
    \end{subfigure}

    \caption{Posterior distributions of latent ability regression coefficients. Histograms show posterior draws; orange solid lines denote posterior means, orange dotted lines denote 95\% credible intervals, and black dashed lines mark zero.}
    \Description{Three posterior histograms summarize the population-level coefficients. Average expected skill change is clearly positive, the coefficient on initial ability is clearly negative, and the coefficient on solo share is positive, with its lower credible bound remaining slightly above zero.}
    \label{fig:alpha_posterior_histograms}
\end{figure}

The posterior distributions of the three population-level coefficients in the latent ability model are shown in Figure~\ref{fig:alpha_posterior_histograms}.   
The posterior distribution of the average expected latent skill change, $\alpha_0$, is positive: for a participant with average initial latent ability and average solo share, the expected latent ability change is $\alpha_0=0.668$, with 95\% CI $[0.485,0.851]$. Because latent ability is measured jointly through accuracy and response time, a positive gain corresponds to improved accuracy and/or faster responses in the post-AI assessment. Figure~\ref{fig:alpha_theta_hist} shows that the coefficient $\alpha_{\theta}$ linking latent skill change to initial ability is strongly negative, with an expected posterior value of $-0.531$ and 95\% CI $[-0.756,-0.307]$, with posterior probability $P(\alpha_{\theta}<0)=1.000$. This indicates that participants with lower initial latent ability tend to show larger gains, consistent with greater room for improvement among initially lower-performing participants. 

Importantly, Figure~\ref{fig:alpha_solo_hist} shows that solo share is positively associated with latent skill change. The posterior mean of $\alpha_{\mathrm{solo}}$ is $0.125$, with 95\% CI $[0.002,0.246]$ and posterior probability $P(\alpha_{\mathrm{solo}}>0)=0.977$. Thus, participants who spent more time independently solving problems  (without AI assistance) during Phase~2 exhibited larger gains in latent ability from the pre-AI to the post-AI assessment. 
To interpret the magnitude of this association, we compute a model-implied contrast between two otherwise average participants whose Phase~2 solo share differs by one minute out of the 20-minute AI-access phase. Relative to the observed population mean value of solo share, this one-minute difference corresponds to a 1.5\% increase in predicted Phase~3 accuracy and a 2.5\% reduction in predicted response time.
Together, these results suggest that, in this context, the key behavioral distinction is not how often participants use AI, but how much independent reasoning they preserve while using it. Participants who preserve more independent reasoning show greater short-term skill gains, consistent with weaker skill development when AI assistance is accompanied by greater cognitive offloading.

As an alternative specification, we also fit a model that replaces solo share with AI usage, measured as the total number of assistance requests during Phase~2 (see details in Supplementary Material~\ref{app:usage_only_model}). %A.5). 
We find that AI usage is not associated with latent skill change after accounting for initial ability, with a posterior for $\alpha_{\mathrm{usage}}$ centered at a posterior mean of $0.0004$,  95\% CI $[-0.122,0.124]$, and $P(\alpha_{\mathrm{usage}}>0)=0.501$. Thus, in this setting, AI request frequency alone does not appear to explain latent skill change. The results are consistent with the hypothesis that time spent on independent reasoning is more predictive of skill development than simple frequency of AI use.

We further compare the main solo-share model discussed above against three alternative specifications: a solo-share model with direct AI-cost condition controls, a usage-only model that replaces solo share with AI usage (the count of the total number of AI requests), and a constant-change baseline in which all participants share the same expected latent skill change, i.e., $\delta_i=\alpha_0$, with no participant-level effects for latent skill change. We evaluate the four specifications using 5-fold cross-validation at the participant level. In each fold, the model is fit on four-fifths of participants and used to predict Phase~3 accuracy and log response time for the remaining participants based on their Phase~1 performance and the Phase~2 predictors included in the corresponding model. 
We evaluate predictive performance using the total held-out log-likelihood, where higher values indicate better overall predictive fit, and the mean-squared error (MSE) for standardized Phase~3 accuracy and log response time, where lower values indicate better point prediction. Table~\ref{tab:model_eval_summary} summarizes the results. 
The solo-share model achieves the best predictive performance across all three metrics, with the highest held-out log-likelihood and the lowest MSE for both accuracy and log response time. 
Additional details of the evaluation procedure are provided in Supplementary Material~\ref{app:model_eval}.% A.6. 

\begin{table}[tb]
\centering
\caption{Five-fold held-out prediction of Phase~3 unassisted performance across Bayesian latent-ability models. The main model uses Phase~2 solo share to predict latent skill change; the condition model adds AI-cost indicators; the AI-usage model replaces solo share with total AI requests; and the constant-change baseline assumes identical expected skill change across participants. Test LogL is the total held-out log predictive likelihood (higher is better). MSE is the mean squared prediction error for standardized Phase~3 accuracy and log response time (lower is better). }
\label{tab:model_eval_summary}

\small
\setlength{\tabcolsep}{3pt}

\begin{tabular}{@{}p{0.50\columnwidth}ccc@{}}
\toprule
\textbf{Model} 
& \shortstack{\textbf{Test}\\\textbf{LogL}$\uparrow$}
& \shortstack{\textbf{Accuracy}\\\textbf{MSE}$\downarrow$}
& \shortstack{\textbf{Log RT}\\\textbf{MSE}$\downarrow$} \\
\midrule
Solo share (main)
& \textbf{-329.94} & \textbf{0.809} & \textbf{0.850} \\

Solo share + AI-cost conditions
& -331.45 & 0.830 & 0.854 \\

AI usage
& -332.96 & 0.831 & 0.882 \\

Constant change (baseline)
& -339.02 & 0.893 & 0.915 \\
\bottomrule
\end{tabular}
\end{table}

\section{Discussion and Conclusion}

This study examines the role of AI assistance in short-term, task-specific skill development.
In a controlled three-phase logic-puzzle experiment, we find that lower-cost AI requests result in increased AI use, and that participants who request AI assistance perform worse after AI assistance is removed. Our Bayesian latent ability model further suggests that skill development is more closely associated with the amount of independent reasoning participants preserve during the AI-access phase than with AI request frequency itself. These findings contribute to the growing literature on AI-enabled deskilling, cognitive offloading, and learning. Prior research shows that external support can improve immediate performance while reducing the cognitive effort needed for later memory, reasoning, or skill formation \citep{risko2016cognitive,grinschgl2021consequences,sparrow2011google}. More recent studies similarly raise concerns that AI assistance can reduce persistence, weaken conceptual understanding, or encourage cognitive disengagement \citep{shen2026ai,kosmyna2025your,dizon2026assessing}. Our results suggest that, in this setting, the central issue is not necessarily AI use per se, but how much independent thinking participants preserve while AI assistance is available. 
The practical implication is that AI systems should be designed not only to improve immediate performance, but also to preserve users' active engagement. Interfaces that encourage an initial attempt, delay assistance, provide partial hints, or prompt reflection may help users benefit from AI without fully offloading the independent reasoning process.
 
This study also points to several directions for future work. Our experiment uses a simulated AI assistant and focuses on short-term skill development in a controlled logic-puzzle task, which helps isolate the relationship between AI use, independent reasoning, and subsequent unassisted performance.  Future studies could examine whether similar patterns persist over longer time horizons, in larger and more diverse samples, and in more realistic domains such as education, professional training, and other high-stakes settings. It would also be useful to study how variation in AI accuracy, explanation quality, and interaction design affects both immediate performance and longer-term skill development. In addition, because the Bayesian model uses participant-level phase aggregates rather than problem-level responses, it does not separately identify residual differences in problem difficulty or problem exposure, providing another potential dimension for future studies.

In conclusion, our work suggests that weaker short-term skill development is associated with greater AI use, particularly when that use involves less independent reasoning. In the context of skill development, designing AI systems that support rather than replace human thinking may therefore be important for improving not only immediate task performance but also subsequent unassisted performance.

%%
% The acknowledgments section is defined using the "acks" environment
% (and NOT an unnumbered section). This ensures the proper
% identification of the section in the article metadata, and the
% consistent spelling of the heading.
\begin{acks}
We thank the reviewers for their valuable feedback that helped improve this paper. This work was supported by the National Science Foundation under awards NSF 2505006 and NSF IIS-2046873, by the Hasso Plattner Institute (HPI) Research Center in Machine Learning and Data Science at UCI, and by funding support from SAP.
\end{acks}

\bibliographystyle{ACM-Reference-Format}
\bibliography{reference}

\newpage
\appendix
\section{Bayesian Modeling and Inference Details}
\label{app:baysian_details}

\subsection{Prior specification}
\label{app:prior_specification}

We use weakly informative priors for all parameters in the Bayesian latent ability model. Because the observed performance measures and predictors are standardized before model fitting, these priors place coefficients on a common scale. For latent initial ability, we use a standard normal prior, $\theta_{i1} \sim \mathcal{N}(0,1)$. For the latent skill-change component, we assign normal priors to the population-level regression coefficients, $\alpha_0,\alpha_{\theta},\alpha_{\mathrm{solo}} \sim \mathcal{N}(0,1)$. We place an exponential prior on the residual skill-change scale, $\sigma_{\delta} \sim \mathrm{Exponential}(2)$. For the measurement model, we use normal priors for the accuracy and response-time intercepts, $\mu_A,\mu_R \sim \mathcal{N}(0,1)$, and exponential priors for the residual standard deviations, $\sigma_A,\sigma_R \sim \mathrm{Exponential}(1)$. These exponential priors restrict scale parameters to positive values and provide weak regularization toward smaller residual variation, consistent with common recommendations for priors on scale parameters in hierarchical Bayesian models \citep{gelman2006prior,stan_prior_recommendations}.

\subsection{MCMC inference}
\label{app:mcmc_inference}

We fit the Bayesian latent ability model in Stan through the Python interface CmdStanPy, using the No-U-Turn Sampler (NUTS), an adaptive Hamiltonian Monte Carlo algorithm \citep{hoffman2014no, stan_users_guide}. For the main specification, we run four chains, each with 1,000 warmup iterations and 2,000 post-warmup sampling iterations, yielding 8,000 posterior draws in total. We set the target acceptance probability to 0.99 and the maximum tree depth to 12.

The model is estimated using a non-centered parameterization for participant-specific latent skill changes, which improves sampling efficiency in hierarchical models by separating the standard-normal skill-change residuals from the residual skill-change scale. Posterior summaries are computed empirically from the retained MCMC draws. Specifically, posterior means are computed as the average of posterior draws, 95\% credible intervals are computed using the 2.5th and 97.5th posterior percentiles, and posterior probabilities such as $P(\alpha_{\mathrm{solo}}>0)$ are computed as the fraction of posterior draws satisfying the corresponding inequality. Participant-level summaries of latent initial ability, post-AI ability, and latent skill change are computed in the same way from the posterior draws of $\theta_{i1}$, $\theta_{i3}$, and $\delta_i$.

\subsection{Model Diagnostics and Predictive Checks}
\label{app:model_checks}

\paragraph{MCMC convergence.}
We assess MCMC convergence using standard sampling diagnostics. For the main solo-share model, all reported $\hat{R}$ values are below 1.01, and the rank-normalized effective sample sizes are satisfactory. We observe no divergent transitions, and the tree-depth and E-BFMI diagnostics indicate no sampling problems. Together, these diagnostics indicate satisfactory convergence and mixing of the posterior samples.

\paragraph{Posterior predictive checks.}
We assess posterior predictive fit by comparing observed Phase~1 and Phase~3 accuracy and log response-time distributions with replicated outcomes from the posterior predictive distribution. The observed means of all four measures and the standard deviations of three of the four fall within their corresponding 95\% posterior predictive intervals; the model somewhat overpredicts the dispersion of Phase~1 log response time. Overall, the model reproduces the central tendencies and most of the observed dispersion well.

\paragraph{Uncertainty in cross-validation comparisons.}
In addition to the aggregate held-out results reported in Table~1,
%\ref{tab:model_eval_summary}, 
we examine uncertainty in predictive performance across held-out participants. The mean held-out log predictive density per participant (standard error) is $-2.661$ ($0.090$) for the main solo-share model, $-2.673$ ($0.090$) for the solo-share model with AI-cost condition indicators, $-2.685$ ($0.090$) for the AI-usage model, and $-2.734$ ($0.088$) for the constant-change baseline. Thus, although the solo-share model has the highest held-out predictive performance, the differences among the specifications are modest. The cross-validation results therefore provide additional predictive evidence in favor of the solo-share specification.

\subsection{Alternative specification with solo share and AI-cost condition controls}
\label{app:condition_controls}

As an alternative specification to the main solo-share model, we allow the assigned AI-cost condition to enter the Bayesian latent ability model directly in addition to solo share. This specification tests whether the randomized condition explains additional variation in latent skill change after accounting for initial ability and solo share. We use the no-AI condition as the reference group and define two indicator variables: $\mathrm{LowCost}_i=1$ for participants assigned to the low-cost AI condition, and $\mathrm{HighCost}_i=1$ for participants assigned to the high-cost AI condition. The latent skill-change component is then
\begin{align*}
\delta_i \sim
\mathcal{N}\left(
\alpha_0
+ \alpha_{\theta}\theta_{i1}
+ \alpha_{\mathrm{low}}\mathrm{LowCost}_i
+ \alpha_{\mathrm{high}}\mathrm{HighCost}_i
+ \alpha_{\mathrm{solo}}\mathrm{Solo}_{i},
\ \sigma_{\delta}
\right).
\end{align*}
The remaining measurement model is unchanged from the main specification. Thus, Phase~1 and Phase~3 accuracy and log response time are modeled as noisy measurements of latent ability, and post-AI ability is defined as $\theta_{i3}=\theta_{i1}+\delta_i$.

The posterior estimates show that adding direct condition effects does not provide clear additional explanatory information. The coefficient for the low-cost AI condition is negative but highly uncertain, $\alpha_{\mathrm{low}}=-0.084$, 95\% CI $[-0.399,0.225]$. The coefficient for the high-cost AI condition is positive but also highly uncertain, $\alpha_{\mathrm{high}}=0.062$, 95\% CI $[-0.238,0.359]$. Both credible intervals contain zero. In contrast, the initial-ability coefficient remains strongly negative, $\alpha_{\theta}=-0.531$, 95\% CI $[-0.762,-0.304]$, consistent with the main model. The solo-share coefficient remains positive but is somewhat attenuated relative to the main specification, $\alpha_{\mathrm{solo}}=0.114$, 95\% CI $[-0.017,0.245]$.

Overall, this alternative specification does not indicate that the assigned AI-cost condition has an additional direct association with latent skill change once initial ability and solo share are included. This supports the main specification, which focuses on preserved independent reasoning during Phase~2 rather than modeling assigned condition as a direct predictor of skill development.

\subsection{Alternative specification with AI usage}
\label{app:usage_only_model}

As an alternative specification, we replace solo share with AI usage in the Bayesian latent ability model. This specification tests whether the total number of AI assistance requests explains latent skill change, rather than the amount of independent reasoning preserved during Phase~2. Let $\mathrm{Usage}_{i}$ denote standardized AI usage, measured as the total number of assistance requests made by participant $i$ during Phase~2. The latent skill-change component is then
\begin{align*}
\delta_i \sim
\mathcal{N}\left(
\alpha_0
+ \alpha_{\theta}\theta_{i1}
+ \alpha_{\mathrm{usage}}\mathrm{Usage}_{i},
\ \sigma_{\delta}
\right).
\end{align*}
The remaining measurement model is unchanged from the main specification. Phase~1 and Phase~3 accuracy and log response time are modeled as noisy measurements of latent ability, and post-AI ability is defined as $\theta_{i3}=\theta_{i1}+\delta_i$.

The posterior estimates show that AI usage does not explain latent skill change. The coefficient on AI usage is essentially zero, $\alpha_{\mathrm{usage}}=0.0004$, 95\% CI $[-0.122,0.124]$, with $P(\alpha_{\mathrm{usage}}>0)=0.501$. Thus, higher request frequency is not associated with either higher or lower latent skill change after accounting for initial ability. The initial-ability coefficient remains negative, $\alpha_{\theta}=-0.495$, 95\% CI $[-0.725,-0.268]$, consistent with the main specification and indicating that participants with lower initial latent ability tended to show larger skill gains.

Overall, this alternative specification suggests that latent skill development is not explained by AI request frequency itself. Together with the main solo-share model, this supports the interpretation that the relevant behavioral signal is the extent to which participants preserved independent reasoning during Phase~2, rather than how often they requested AI assistance.

\subsection{Held-out model evaluation}
\label{app:model_eval}

We compare four Bayesian latent ability model specifications. The first is the main solo-share model described in Section~\ref{subsec:bayesian_model}, 
in which latent skill change depends on initial latent ability and Phase~2 solo share. The second adds direct AI-cost condition controls to the solo-share model, as described in Supplementary Material~\ref{app:condition_controls}. The third is the AI-usage model in Supplementary Material~\ref{app:usage_only_model}, which replaces solo share with the total number of AI requests. The fourth is a constant-change baseline, in which all participants share the same expected latent skill change, $\delta_i=\alpha_0$, so post-AI ability differs only through baseline latent ability and measurement noise.

We evaluate these specifications using 5-fold held-out prediction of Phase~3 outcomes. In each fold, the model is fit using only the four-fifths of participants assigned to the training fold. For these training participants, the model observes the Phase~1 performance measures, Phase~3 outcomes, and any behavioral predictors used by the corresponding specification. The remaining one-fifth of participants are excluded from model fitting entirely. After fitting, we use only the held-out participants' prediction inputs to compute posterior predictive distributions for their Phase~3 outcomes. These inputs are Phase~1 performance and solo share for the main solo-share model; Phase~1 performance, solo share, and condition indicators for the condition-control model; Phase~1 performance and AI usage for the usage-only model; and Phase~1 performance only for the constant-change baseline.

The primary evaluation metric is held-out log likelihood, which measures out-of-sample predictive accuracy by evaluating the posterior predictive probability assigned to held-out observations. For each held-out participant, we compute the log predictive probability of the observed Phase~3 accuracy and log response time under the posterior predictive distribution, averaging over posterior draws from the model fit on the training fold. We then sum this quantity across held-out participants and folds, with higher values indicating better predictive fit. We also report held-out mean squared error (MSE), computed by comparing posterior mean predictions with the observed held-out Phase~3 accuracy and log response time; lower MSE values indicate better point prediction.

As reported in Table~\ref{tab:model_eval_summary}, 
the main solo-share model achieves the highest held-out log likelihood and the lowest held-out MSE for both Phase~3 accuracy and log response time. Thus, the held-out results favor the solo-share model on overall predictive fit.

\section{Experiment Details}
\subsection{Post-Study Survey}
\label{appendix:post-study_survey}
After completing the three-phase study, participants were asked to fill out a short post-study survey. This survey was originally used in the pilot study to gather feedback and adjust the game settings, and we kept it in the main experiments for consistency. In the analyses reported in this paper, we only use responses from participants in AI conditions regarding whether they reported using AI assistance, comparing these with actual usage logs to identify inconsistent behavior as a filtering condition. No other survey responses were used in our analyses. 

The survey content is shown below.

Questions with 5-point Likert scale response options: \textit{Strongly Disagree}, \textit{Disagree}, \textit{Neutral}, \textit{Agree}, and \textit{Strongly Agree}:

\begin{itemize}[noitemsep, topsep=0pt]
    \item I found the game to be boring.
    \item I felt confident in my performance throughout the game.
    \item I put in considerable effort to achieve my level of performance.
    \item I developed a strategy to improve my performance during the game.
    \item I find seeing the correct solution for each problem helps me learn.
    \item I’ve had prior experience with a similar task.
\end{itemize}

\noindent Two optional free-response questions:
\begin{itemize}[noitemsep, topsep=0pt]
    \item Did you notice any hidden rules during the game? If so, please
    describe them. Additionally, please share the strategy you used while solving the puzzles.
    \item Please share your overall experience with the game and any suggestions for improvement.
\end{itemize}

\noindent For participants in AI assistance conditions, we further asked:
\begin{itemize}[noitemsep, topsep=0pt]
    \item Rate the AI assistant's accuracy (0–100, or select Not Applicable).
    \item The AI assistant was helpful during the experiment (5-point Likert scale, with an additional choice \textit{Not Applicable}).
    \item When did you typically use the AI assistant during the experiment? For example: Right after seeing the problem, after attempting a submission, to check your answer, etc. (optional free response question)
\end{itemize}

\end{document}